\documentclass[10pt,twocolumn,letterpaper]{article}

\usepackage{iccv}
\usepackage{times}
\usepackage{epsfig}
\usepackage{graphicx}
\usepackage{amsmath}
\usepackage{amssymb}
\usepackage{xcolor}
\usepackage{diagbox}
\usepackage{multirow}
\usepackage{booktabs}
\usepackage{bm}
\usepackage{arydshln}
\usepackage{pifont}%
\newcommand{\cmark}{\ding{51}}%
\newcommand{\xmark}{\ding{55}}%

\usepackage[accsupp]{axessibility}  %

\usepackage[pagebackref=true,breaklinks=true,letterpaper=true,colorlinks,bookmarks=false]{hyperref}
\usepackage[capitalize]{cleveref}

\iccvfinalcopy %

\def\iccvPaperID{3966} %

\ificcvfinal\fi

\newcommand{\method}{TMC}
\newcommand{\methodensemble}{TME}
\newcommand{\figref}[1]{Fig.~\ref{#1}}
\newcommand{\tabref}[1]{Table~\ref{#1}}
\newcommand{\secref}[1]{Sec.~\ref{#1}}

\begin{document}

\title{
Tangent Model Composition for Ensembling and Continual Fine-tuning
}

\author{Tian Yu Liu\\
University of California, Los Angeles\\
{\tt\small tianyu@cs.ucla.edu}
\and
Stefano Soatto\\
University of California, Los Angeles\\
{\tt\small soatto@ucla.edu}
}

\maketitle
\ificcvfinal\thispagestyle{empty}\fi

\begin{abstract}
Tangent Model Composition (TMC) is a method to combine component models independently fine-tuned around a pre-trained point. Component models are tangent vectors to the pre-trained model that can be added, scaled, or subtracted to support incremental learning, ensembling, or unlearning. Component models are composed at inference time via scalar combination, reducing the cost of ensembling to that of a single model. TMC improves accuracy by 4.2\% compared to ensembling non-linearly fine-tuned models at a 2.5$\times$ to 10$\times$ reduction of inference cost, growing linearly with the number of component models. Each component model can be forgotten at zero cost, with no residual effect on the resulting inference.  When used for continual fine-tuning, TMC is not constrained by sequential bias and can be executed in parallel on federated data. TMC outperforms recently published continual fine-tuning methods almost uniformly on each setting -- task-incremental, class-incremental, and data-incremental -- on a total of 13 experiments across 3 benchmark datasets, despite not using any replay buffer.  TMC is designed for composing models that are local to a pre-trained embedding, but could be extended to more general settings. The code is available at: \url{https://github.com/tianyu139/tangent-model-composition}
\end{abstract}

\section{Introduction}
\begin{figure*}[ht]
    \centering
    \includegraphics[width=0.9\linewidth]{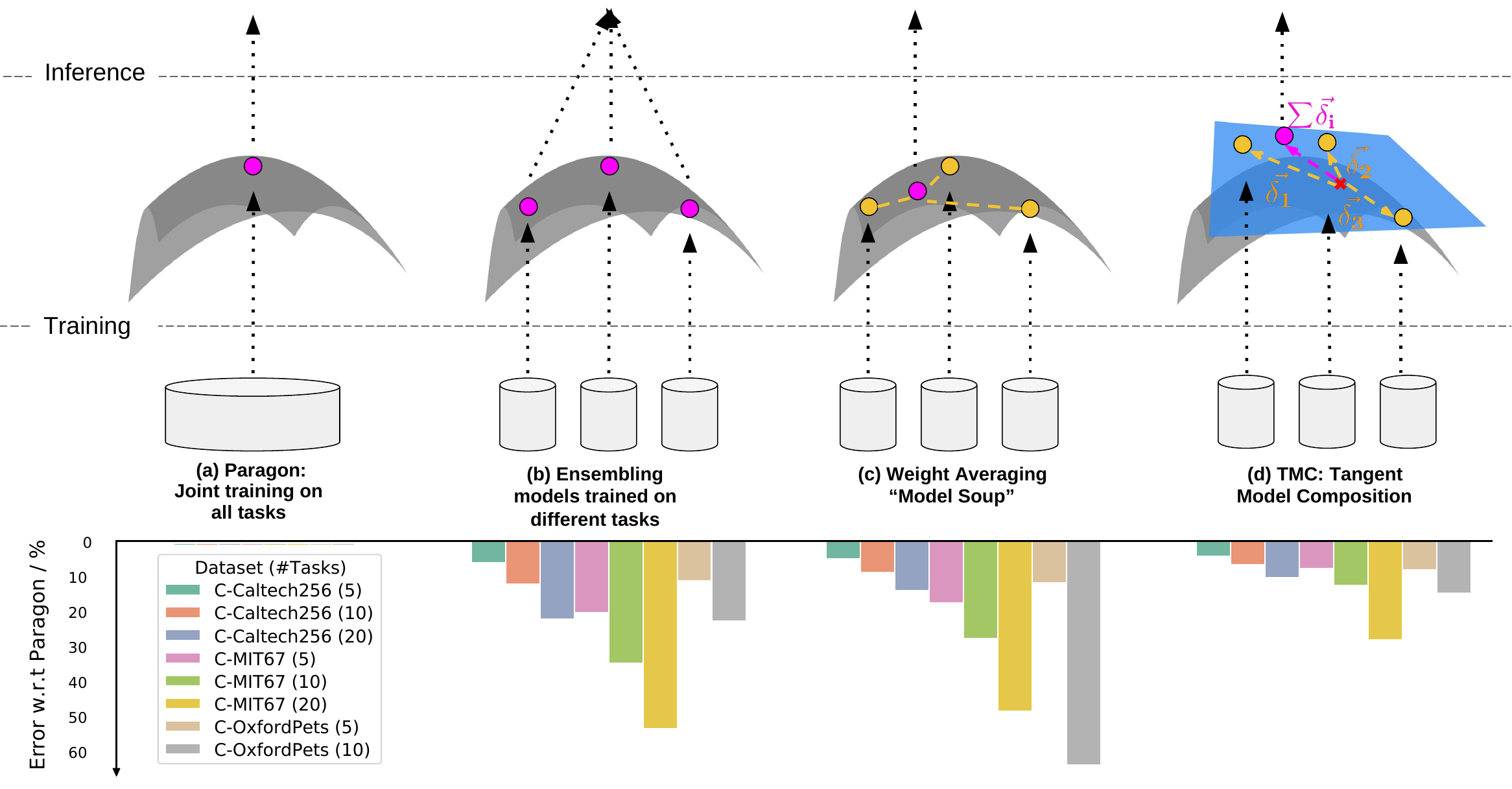}
    \caption{{\bf Composition of Fine-tuned Models:}  Purple indicates final models used for inference, yellow indicates specialist models fine-tuned on individual tasks. The paragon (a) is a model trained jointly on all tasks. However, in continual learning, different tasks are instantiated at different times, an re-training on their union is impractical. Ensembling (b) combines the output of different models trained separately on each task. Weight averaging (c) yields a  ``model soup'' of fine-tuned non-linear models, which improves inference time; Tanget Model Composition (d) linearly composes models fine-tuned on the tangent space of a pre-trained model.}
    \label{fig:main}
\end{figure*}
Compositionality has long been considered central to cognition  \cite{bienenstock1998compositionality}, possibly a reflection of compositionality {\em in neural activity} \cite{tsao2003faces}.
After decades of unsuccessful attempts to design compositional representations, deep neural networks trained on language data are beginning to exhibit emergent compositionality at scale, by capturing the compositional structure of the data. These models are now being used for visual inference, which has rekindled interest in the study of compositionality of visual representations.  But beyond the data, any compositional structure that may be latent in the model is not directly accessible: One cannot simply ``compose'' weights or inner activations and expect meaningful outcomes. The computational architecture of Transformers \cite{vaswani2017attention} has been leveraged extensively to co-opt the compositional structure of data through prompts or tokens \cite{lester2021power,wang2022dualprompt}, but still the activations of trained models do not appear to be meaningfully composable. 
Compositionality of neural activity would allow one to combine activations from different models to capture novel concepts, or incorporate knowledge from different data without having to re-train or fine-tune the core models. This would enable open-universe classification and, more generally, combinatorial expansion of the hypothesis space. Continual learning could be performed simply by composing models trained on different data.

In this paper, we explore the simplest form of compositionality, that is {\em linear combination}. We leverage recent results on the linearization of deep neural networks around a pre-trained point, that can be trained by solving a convex optimization problem and yet perform on-par with non-linear fine-tuning \cite{achille2021lqf}. This suggests that the tangent space at pre-trained models may be used to linearly compose neural activations, or equivalently compose different models trained or fine-tuned on different datasets and/or for different tasks.

We show that different models obtained from the linearization around a pre-trained point can be composed, combined, rescaled, and forgotten (``unlearning''), simply by scalar combinations. This fact can be used to perform ensembling at the inference cost of a single model (\secref{sec:method}). It can also be used for continual fine-tuning, with each component model trained independently and in parallel, if necessary on federated data that can therefore be easily forgotten if needed.

Linear combination is not viable for general concept composition nor arbitrary multi-task learning. In \secref{sec:discussion}, we discuss limitations of our approach, which can only meaningfully compose models that are ``local'' to a pre-trained embedding. If there are models trained on tasks that are {\em far} in representation space, or even {\em antagonistic}, they are likely to live on different tangent planes, making linear combination inappropriate. One such example, described in the appendix, concerns models trained on real images (ImageNet) and clip art or sketches (DomainNet).  
Nonetheless, in more common settings, \method{} is competitive with general forms of ensembling such as averaging activations or logits of non-linearly fine-tuned models, and with more general forms of continual learning, including methods that use a replay buffer. For broader task coverage, one could use \method{} to construct a {\em tangent bundle of models} around different pre-trained points, akin to a ``tangent model zoo.'' This concept is not too dissimilar from the architecture of some Foundation Models with a shared backbone and multiple distinct ``heavy heads'' \cite{wang2022ofa}, but far more compact, modular, efficient, interpretable, and easy to work with.

In the next section we place our contributions in the context of prior art, and in the following one we describe our method in more detail. \secref{sec:experiments} summarizes empirical evidence in support of our approach, and finally \secref{sec:discussion} discusses its main limitations.

\section{Related work}
\textbf{Compositionality:}
Compositionality has been studied for decades as a means to expand the representative power of trained models, but has received increasing impetus in recent years thanks to large Transformer-based models that can be used with adaptable prompts. Prompt-tuning \cite{lester2021power} has become commonplace to  adapt large pre-trained models for specific downstream tasks. Composition of prompts has also been explored in \cite{wang2022learning,wang2022dualprompt} for continual learning. More general composition of parameters of trained deep networks has been explored for various purposes such as improving optimization and generalization, as we describe next.

\textbf{Deep network linearization: } Deep networks are linearized using the first-order Taylor expansion around a pre-trained weight \cite{achille2021lqf,golatkar2021mixed,mu2020gradients,shon2022dlcft,liu2022integral}. \cite{achille2021lqf} finetunes linearized networks modified to use Leaky-ReLU with gradient pre-conditioning to achieve similar performances to non-linear fine-tuning.
We note that TMC is not just a matter of applying linearized networks to different datasets and averaging, since different scales would lead to imbalance as we describe in \secref{sec:method-rsl}. Furthermore, unlike \cite{achille2021lqf,shon2022dlcft}, we do not require Leaky-ReLU nor gradient pre-conditioning. 

\textbf{Averaging model weights:} Weight averaging has been used in deep learning to improve generalization \cite{izmailov2018averaging,garipov2018loss}, improve pre-training \cite{choshen2022fusing,matena2021merging}, perform distributed fine-tuning \cite{wortsman2022fi}, and increase robustness against distribution shift \cite{wortsman2022robust}. \cite{izmailov2018averaging} averages points along the trajectory of SGD to find flatter minima, \cite{von2020neural} averages ``late-phase weights" obtained from the later stages of SGD. \cite{ilharco2022patching} uses weight interpolation for model patching, and \cite{wortsman2022model} averages weights of models fine-tuned using different hyperparameters from pre-trained models to improve generalization. 

While weight averaging techniques for deep networks are related to ensembles \cite{dietterich2000ensemble,gontijo2021no}, in our case the two coincide since we operate in representation spaces that are linear in the parameters. We leverage this property to  train on disjoint subsets of data, and show that they can outperform existing weight averaging and ensembling techniques.

\textbf{Continual learning/fine-tuning:} Continual learning \cite{parisi2019continual} aims to adapt models to new tasks and data without degrading performance in previously trained tasks. Continual fine-tuning \cite{shon2022dlcft,boschini2022transfer} additionally requires minimal forgetting of a pre-training objective to ensure all tasks can benefit from it. 
Existing approaches can be broadly classified into parameter isolation, regularization, architecture, and replay-based.

Parameter isolation methods \cite{liu2022integral,aljundi2017expert,rusu2016progressive} operate in the network architecture space by allocating each task a different set of parameters, using techniques such as network pruning \cite{mallya2018packnet} to reduce additional memory requirements.

Regularization-based methods \cite{li2017learning,aljundi2018memory,kirkpatrick2017overcoming,nguyen2017variational,lee2020continual} incorporate additional terms into the training objective to prevent catastrophic forgetting. \cite{kirkpatrick2017overcoming} penalizes the distance between weights of the previous and current task, and \cite{ritter2018online} penalizes changes in weights using Hessian approximations.

Experience replay methods \cite{riemer2018learning,aljundi2019gradient,liu2022integral,chaudhry2018efficient,rebuffi2017icarl,shon2022dlcft,buzzega2020dark,buzzega2021rethinking,kirkpatrick2017overcoming,wang2022learning,boschini2022transfer,cha2021co2l,caccia2022new} assume the availability of a memory buffer to store samples from previous tasks. \cite{shon2022dlcft,boschini2022transfer} applies experience replay to continual fine-tuning. \cite{shon2022dlcft} shows that linearized models combined with replay and regularization alleviates catastrophic forgetting when fine-tuned, and \cite{boschini2022transfer} combines dark experience replay and an attention-based regularization loss to prevent catastrophic forgetting of the pre-training features, by encouraging similarity of features at each layer to the original pre-trained network.
While replay methods have achieved state-of-the-art performance compared to non-replay ones, they require a sufficiently large memory buffer to store previously seen examplars. 

A further line of work tackles exemplar-free class incremental learning (EFCIL) using regularization \cite{zhu2021class}, distillation \cite{hou2019learning}, and class-prototypes \cite{zhu2021prototype,zhu2022self,petit2023fetril} to incrementally integrate new classes into a fixed-sized network without the memory costs of a replay buffer. 

We show that composition of tangent models trained on individual tasks is a viable method for replay-free continual fine-tuning, and can even outperform recently published replay-based methods. As a side benefit, our method also yields fully parallel training across tasks, and enables zero-cost unlearning. %

\section{Method}
\label{sec:method}
\begin{table}[h]
    \centering
    \caption{\textbf{A Comparison of Composition Methods:} On average across 25 experiments, Tangent Model Ensembling (TME) improves accuracy by 5.1\% over standard ensembling of non-linear fine-tuned models at a 2$\times$ increase of inference cost. Tangent Model Composition (TMC) also improved accuracy, by 4.2\%, while {\em reducing} inference cost by 2.5$\times$ to 10$\times$ in our experiments, and growing linearly with the number of models in the ensemble.  Best results for single and multi model inference are indicated in bold and italic respectively. 
    Joint training paragon accuracy (Original Model/Tangent Model):
    Caltech-256 - $89.17\%$/$86.77\%$, MIT-67 - $77.29\%$/$77.74\%$, OxfordPets - $92.68\%$/$93.68\%$.
  }
    \resizebox{\linewidth}{!}{%
    \setlength{\tabcolsep}{2.3pt}
    \begin{tabular}{ccccccc}
    \toprule
     Dataset & Tasks & Soup & Ens-L  & Ens-SM & \method{} & \methodensemble{}  \\
     \midrule
     \multicolumn{2}{r}{Inference/Memory Cost:} & $\mathcal{O}(1)$ & $\mathcal{O}(T)$ & $\mathcal{O}(T)$ & $\mathcal{O}(1)$ & $\mathcal{O}(T)$\\
         \midrule
\multirow{3}{*}{\shortstack{C-Caltech-256\\\small{\textit{(Class-IL)}}}} & 5  &  $84.00$ & $82.80$ & $83.09$ & $\bf{84.82}$ & $\textit{85.53}$ \\
& 10 &  $80.10$ & $76.85$  & $79.76$ & $\bf{82.37}$  & $\textit{82.58}$ \\
& 20 &  $74.86$ & $66.76$  & $75.47$ &  $\bf{78.66}$ & $\textit{79.30}$ \\
\hdashline
\multirow{3}{*}{\shortstack{C-MIT-67\\\small{\textit{(Class-IL)}}}} & 5 & $59.48$ & $56.79$ & $58.68$ & $\bf{69.43}$ & $\textit{71.72}$ \\
& 10 &  $49.30$   & $42.36$ & $50.52$ &  $\bf{64.53}$ & $\textit{65.72}$  \\
& 20 &  $28.51$  & $23.61$ & $35.62$ & $\bf{48.98}$  & $\textit{54.93}$ \\
\hdashline
\multirow{2}{*}{\shortstack{C-OxfordPets\\\small{\textit{(Class-IL)}}}} & 5 & $80.71$  & $81.14$ &  $81.86$  &  $\bf{84.42}$ & $\textit{85.04}$ \\
& 10 &  $71.35$ & $69.71$ & $77.31$    &  $\bf{77.75}$  & $\textit{79.97}$ \\
\midrule
\multirow{3}{*}{\shortstack{D-Caltech-256\\\small{\textit{(Data-IL)}}}} & 5  &  $\bf{85.96}$ & $86.27$ & $\textit{86.59}$ & $84.86$ & $85.13$ \\
& 10 &  $83.48$ & $84.05$  & $\textit{84.80}$ & $\bf{83.64}$  & $84.16$ \\
& 20 &  $79.64$ & $80.72$  & $\textit{83.17}$ & $\bf{81.99}$ & $82.94$ \\
\hdashline
\multirow{3}{*}{\shortstack{D-MIT-67\\\small{\textit{(Data-IL)}}}} & 5 & $66.00$ & $66.54$ & $67.21$ & $\bf{74.88}$ & $\textit{75.45}$ \\
& 10 &  $58.88$   & $59.20$ & $60.47$ & $\bf{72.41}$ & $\textit{73.41}$  \\
& 20 &  $50.10$  & $51.99$ & $55.20$ & $\bf{70.22}$  & $\textit{72.11}$ \\
\hdashline
\multirow{3}{*}{\shortstack{D-OxfordPets\\\small{\textit{(Data-IL)}}}} & 5 & $90.22$  & $90.83$ & $91.02$ &  $\bf{93.17}$ & $\textit{93.21}$ \\
& 10 &  $87.66$ & $88.42$ & $88.88$ &  $\bf{92.48}$  & $\textit{92.72}$ \\
& 20 &  $85.57$ & $85.97$ & $86.58$ &  $\bf{91.28}$  & $\textit{91.77}$ \\
\midrule
\multirow{3}{*}{\shortstack{C-Caltech-256\\\small{\textit{(Task-IL)}}}} & 5  & $94.01$ & $94.72$ & $\textit{94.84}$ & $\bf{94.33}$ & $94.76$ \\
& 10 & $95.00$ & $95.87$ & $96.04$ & $\bf{96.03}$ & $\textit{96.23}$ \\
& 20 & $95.56$ & $96.53$ & $96.84$ & $\bf{97.22}$ & $\textit{97.38}$ \\
\hdashline
\multirow{3}{*}{\shortstack{C-MIT-67\\\small{\textit{(Task-IL)}}}} & 5 & $83.36$ & $86.24$ & $86.37$ & $\bf{91.27}$ & $\textit{91.79}$ \\
& 10 & $87.19$ & $90.42$ & $90.97$ & $\bf{94.45}$ & $\textit{95.03}$ \\
& 20 & $88.41$ & $93.73$ & $93.88$ & $\bf{97.71}$ & $\textit{98.09}$ \\
\hdashline
\multirow{2}{*}{\shortstack{C-OxfordPets\\\small{\textit{(Task-IL)}}}} & 5 & $97.27$ & $97.27$ & $97.35$ & $\bf{98.57}$ & $\textit{98.74}$\\
& 10 & $97.59$ & $97.92$ & $97.98$ & $\bf{99.05}$ & $\textit{99.13}$ \\
    \bottomrule
    \end{tabular}
    }
    \label{tab:continual-composition}
    \vspace{-3mm}
\end{table}

In continual learning, models are trained on a sequence of tasks indexed by $t=1,\ldots,T$, with  the goal of performing well on all, under three main scenarios \cite{van2019three}: (1) \textit{Task-incremental}, where the task identity $t$ (Task-ID) is known during inference, (2) \textit{Data/Domain-incremental}, where Task-ID is not provided during inference, and (3) the most challenging \textit{Class-incremental}, where Task-ID is unknown during inference and in addition the classes represented in each task are disjoint. 

We assume the setting of continual fine-tuning of a pre-trained model. Naive fine-tuning causes catastrophic forgetting of prior tasks, including the one on which pre-training occurred, as the number of tasks increases \cite{boschini2022transfer}. Thus, only early tasks benefit from pre-training. Therefore, it is important for continual fine-tuning to ensure that all tasks can benefit from information acquired during pre-training.

We introduce a unified method to tackle the main scenarios (1)-(3)  by composing separately-trained tangent component models. We show that each task takes full advantage of pre-training, and catastrophic forgetting of prior tasks is averted since information from each previous task is encoded separately and only fused for inference.

\subsection{Model Linearization}
The effectiveness of model aggregation in weight space hinges on whether the models lie in within the same small-loss training basin of convergence. It is known that weights of different models fine-tuned on the same dataset can be connected by linear paths with constant loss (``linear mode connectivity'') \cite{frankle2020linear}, but this does not generally occur for models fine-tuned on disjoint tasks/datasets, which is the case of interest to us. Therefore, we convexify the loss landscape locally by approximating the function represented by a deep (non-linear) network with its first-order Taylor approximation around the pre-trained weights. For any deep network $f$ and initial weights $w \in \mathbb{R}^m$, we denote with $\mathcal{H}(w)$ the set of all such models tangent to $f_w$ given by:
\begin{align}
    \mathcal{H}(w) = \{h_{\delta}(\cdot) \triangleq  f_w(\cdot) + \nabla_w f_w(\cdot)\cdot\delta \}.
\end{align}
Since the tangent model $h_\delta$ is linear in the parameters $\delta$, training it with standard losses such as mean-squared error (MSE) or empirical cross-entropy yields a convex loss landscape. In order to compose models, leveraging Jensen's inequality, we are guaranteed that for any dataset $\mathcal{D}$:
\begin{align}
    L(\mathcal{D}, \sum_i^T h_{\theta_i \delta_i}) \leq \sum_i \theta_i L(\mathcal{D}, h_{\delta_i}), \quad \sum_i \theta_i = 1.
\end{align}
In other words, the loss resulting from a convex combination of tangent models in weight space will always be lower than or equal to the convex combination of their individual losses. This is particularly useful when combining models trained to some constant (possibly zero) loss, since it ensures that the composed model will have loss lower or equal to the loss of the component models.

While tangent models might seem computationally expensive to train, we note that the Jacobian vector product ${\nabla_w f_w(\cdot)\cdot\delta}$ can be computed in a single forward pass as shown by \cite{pearlmutter1994fast,mu2020gradients,achille2021lqf}. This reduces the computational cost to that of a forward pass through the original network. Also note that while the tangent model has twice the degrees of freedom of the original model, fixing the pre-trained model at $w$ results in only a convex optimization in the remaining parameters, as many as the original model. 

In the context of fine-tuning, $w$ is initialized by pre-training on large datasets such as ImageNet \cite{deng2009imagenet}. In the rest of this paper, we use the short-form notation $\mathcal{H}$ and $h$ for $\mathcal{H}(w)$ and $h_{\delta}$ respectively.

\subsection{Tangent Model Composition for Replay-free Continual Fine-tuning}
\begin{figure}
    \centering
    \includegraphics[width=0.95\linewidth]{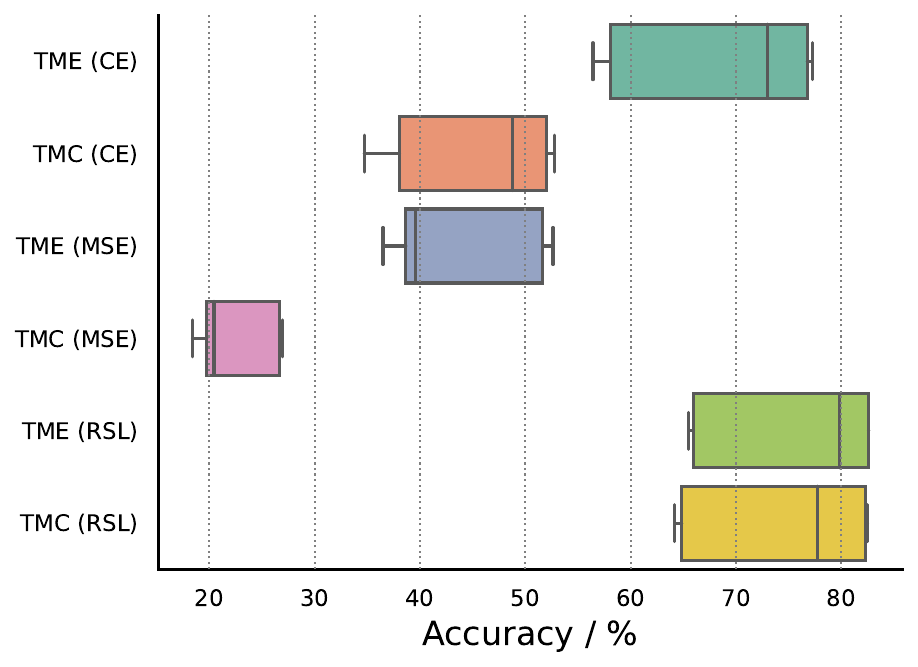}
    \caption{Accuracy plot over models trained on C-Caltech-256, C-MIT-67, and C-Oxford Pets, split into 10 tasks each. Soft-max ensembling (\methodensemble{}) consistently yields the best results compared to logit ensembling (\method{}) for each loss function. 
    Composition of component models trained with the Rescaled Square Loss (RSL) outperforms those trained with empirical cross-entropy (CE) or MSE loss, trading off only $1.2\%$ accuracy against \methodensemble{} to reduce inference time and memory requirement from $\mathcal{O}(T)$ to $\mathcal{O}(1)$.}
    \label{fig:compose-plot}
\end{figure}

\begin{table*}[t]
\centering
\caption{{\bf Class-Incremental Learning:} Comparison against existing methods using an ImageNet pre-trained ResNet-50. \method{} is almost uniformly better than all replay-based and replay-free methods tested, outperforming the best replay-free (RF) method by $14.48\%$ and even the best replay-based method \cite{boschini2022transfer} by $1.53\%$.}
\begin{tabular}{lccc|ccc}
\toprule
Method / Buffer      & \multicolumn{3}{c}{C-Caltech-256} & \multicolumn{3}{c}{C-MIT-67} \\
 \cmidrule(lr){2-4} \cmidrule(lr){5-7}  
\multicolumn{1}{r}{Tasks}       & 5        & 10        & 20       & 5      & 10      & 20     \\
\midrule
ER \cite{robins1995catastrophic} / 300  & $70.14$\tiny{$\pm0.22$} & $65.75$\tiny{$\pm0.92$}  &   $65.49$\tiny{$0.60$}  &  $54.61$\tiny{$\pm0.41$} & $49.57$\tiny{$\pm1.46$}  & $45.33$\tiny{$1.73$} \\
LWF \cite{li2017learning} / \textbf{None} & $34.55$\tiny{$\pm0.18$} & $22.76$\tiny{$\pm0.17$} & $14.37$\tiny{$\pm0.14$} & $25.14$\tiny{$\pm0.57$} & $13.30$\tiny{$\pm0.49$} & $6.41$\tiny{$\pm0.17$}\\
oEWC \cite{schwarz2018progress} / \textbf{None} & $28.71$\tiny{$\pm1.31$} & $16.32$\tiny{$\pm4.37$} & $15.59$\tiny{$\pm3.83$} & $14.18$\tiny{$\pm0.77$} & $7.12$\tiny{$\pm0.49$} & $4.33$\tiny{$\pm0.16$} \\
DER++ \cite{buzzega2020dark} / 300 & $80.50$\tiny{$\pm0.26$} & $77.84$\tiny{$\pm1.02$} & $74.46$\tiny{$\pm0.21$} &  $64.56$\tiny{$\pm0.55$} &  $59.73$ \tiny{$\pm1.02$} & $50.28$\tiny{$\pm1.27$} \\
Co2L \cite{cha2021co2l} / 300   &  $20.71$\tiny{$\pm2.89$}        & $17.13$\tiny{$\pm2.53$}   &  $14.62$\tiny{$0.61$}  & $19.64$\tiny{$\pm12.93$}  &  $21.21$\tiny{$\pm0.50$}    &  $17.78$\tiny{$\pm1.33$} \\
ER-ACE \cite{caccia2022new} / 300  & $78.47$\tiny{$\pm1.25$} &   $73.85$\tiny{$\pm0.53$}  & $70.87$\tiny{$\pm1.38$} &  $64.94$\tiny{$\pm0.50$}  & $60.25$\tiny{$\pm0.26$}  & $52.48$\tiny{$\pm0.29$} \\
TWF (RF) \cite{boschini2022transfer} / \textbf{None} &  $77.95$\tiny{$\pm0.52$} & $70.67$\tiny{$\pm0.72$} &   $62.79$\tiny{$\pm0.82$} & $56.58$\tiny{$\pm0.62$} &  $45.03$ \tiny{$\pm0.78$}  &  $28.91$\tiny{$\pm0.92$} \\
TWF \cite{boschini2022transfer} / 300 &  $81.34$\tiny{$\pm0.91$} & $77.03$\tiny{$\pm0.53$} &   $72.44$\tiny{$\pm0.42$} & $68.33$\tiny{$\pm0.46$} &  $64.21$ \tiny{$\pm0.65$}  &  $\bm{56.29}$\tiny{$\pm0.68$} \\
\method{} / \textbf{None}    &  $\bm{84.82}$ \tiny{$\pm0.03$}        &           $\bm{82.37}$ \tiny{$\pm0.08$} &  $\bm{78.66}$ \tiny{$\pm 0.13$}   &  $\bm{69.43}$\tiny{$\pm0.21$}    & $\bm{64.53}$\tiny{$\pm0.28$}     &   $48.98$ \tiny{$\pm0.13$}    \\
\bottomrule
\end{tabular}
\label{tab:class-incremental-sota}
\vspace{-3mm}
\end{table*}

Ensembling is an effective strategy for combining weak learners into a strong model \cite{dietterich2000ensemble}. For any finite $F \subseteq \mathbb{A}$, where $\mathbb{A}$ is a set of functions, and ${\lambda = \{\lambda_i\}_{i=1}^{|F|}}$ a set of weights, the logit ensemble operation is given by
\begin{align}
    Ensemble(F, \lambda) = \sum_{i=1}^{|F|} \lambda_i f_i(\cdot)
\end{align}
In the context of continual learning, a simple yet effective approach is to simply train a new model on each task $t$, and ensemble the outputs at inference time. Unfortunately, this has the obvious shortcoming that both storage requirements and inference time grows along with the number of tasks. 

However, consider the case where $\mathbb{A} = \mathcal{H}$. Then for any finite $H \subseteq \mathcal{H}$ and associated weights $\lambda$, we can write
\begin{align}
Ensem&ble(H,\lambda) = \sum_{i=1}^{|H|} \lambda_i h_i(\cdot) \\
 &= f_w(\cdot) + \sum_{i=1}^{|H|} \nabla_w f_w(\cdot)\cdot \lambda_i\delta_i \\
 &= h_{{}_{\sum \lambda_i \delta_i}}(\cdot)
\end{align}
In other words, since $\mathcal{H}$ defines a vector space with addition and scalar multiplication based on $\delta$, any linear combination (or ensemble) of models in $\mathcal{H}$ is equivalent to a single model in $\mathcal{H}$ derived simply by the linear combination over $\delta$.
This reduces the inference time for any ensemble of $T$ tangent models from $\mathcal{O}(T)$ to $\mathcal{O}(1)$. 

In continual learning, it is also important for storage space to not scale with $T$, since the number of tasks can be arbitrarily large. However, since tasks arrive sequentially in the continual learning framework, we can further reduce the memory capacity required for storage of models from $\mathcal{O}(T)$, to $\mathcal{O}(1)$ by using a simple autoregressive model: We train on the first task $t_1$ to produce model $\tilde{h}_1 = h_1$. Then, for any subsequent task $t > 1$, we train a new model  $\tilde{h}_t$, and compose it with $h_{t-1}$ to give $h_t = \lambda_{(t,1)} \tilde{h}_t + \lambda_{(t,2)} h_{t-1} := h_{\delta'}$ where $\delta'=\frac{\lambda_{(t,1)}\tilde{\delta}_t + \lambda_{(t,2)}\delta_{t-1}}{\lambda_{(t,1)} + \lambda_{(t,2)}}$. 
 Since we only require model $h_{t-1}$ to produce $h_t$, all prior models up to and including $h_{t-1}$ can be discarded upon constructing $h_t$, hence reducing memory requirement from $\mathcal{O}(T)$ to $\mathcal{O}(1)$, equal to the size of a single model. In contrast to methods that expands the model at each task, or require a memory buffer of examples from previous tasks, this method does not incur additional storage requirements, nor assume availability of an extra replay buffer for storing past exemplars (\textit{i.e.} our method is replay-free).

\noindent\textbf{Remark 1 (Free Parallelization):} Additionally, for any given convex loss function $L$, the optimization objective
\begin{align}
    \mathop{\min}\limits_{\delta \in \mathbb{R}^m} \sum_{(x,y) \in \mathcal{D}} L(h_{\delta}(x), y) 
\end{align}
is convex and thus converges to a global optimum. This implies that each model $\tilde{h_t}$ can be trained from a standard initialization as opposed to from $h_{t-1}$. Hence, if multiple tasks arrive at the same time, such as when training in a federated manner, training can be parallelized across tasks and composed in a zero-cost, zero-shot manner to yield equivalent results (assuming equivalence of global optima) to sequential training and composition. To the best of our knowledge, apart from \cite{prabhu2020gdumb} which simply stores samples in a replay buffer and trains on them at test time, no other work in continual learning can be parallelized across tasks.

\noindent\textbf{Remark 2 (Free Forgetting/Unlearning): } If we relax memory constraints to store models from each task,  unlearning any specific task $t$ (\textit{e.g.}, private data associated with specific samples) is straightforward and can be done in a zero-cost, zero-shot manner. Since subtraction is well-defined in the vector space $\mathcal{H}$, any task $i$ can be unlearned from the model $h_T$ trained on $T$ tasks by simply subtracting the weights associated to $\tilde{h}_i$:
\begin{align}
    h_{T \setminus i} = h_T - \lambda_{i,1} \prod_{j=i+1}^T \lambda_{(j,2)} \tilde{h}_i
\end{align}

\noindent\textbf{Remark 3 (Unified Continual Fine-tuning): } Our method does not assume any specific task splits, and is hence generalizable across all three forms of continual fine-tuning: task-incremental, class-incremental, and data-incremental.

\subsection{Scale Standarization}
\label{sec:method-rsl}

Training with the cross-entropy loss does not ensure that the weights are comparable in scale. We can study the implications of this for composition on $\mathcal{H}$ through the lens of logits ensembling. 
While logits ensembling has been shown to be effective \cite{wortsman2022model}, \figref{fig:compose-plot} shows that ensembling the normalized soft-max outputs (Tangent Model Ensembling, \methodensemble{}) is uniformly better than logit ensembling/weight averaging via \method{} across various loss functions.
However, \methodensemble{} cannot be represented by a linear combination of perturbations $\delta_i$ and hence incurs $\mathcal{O}(T)$ model storage costs and $\mathcal{O}(T)$ inference time. 

Instead, we propose to induce a waterbed effect similar to soft-max ensembling using standardization. In particular, we train using the Rescaled Square Loss (RSL) \cite{hui2020evaluation}:
\begin{align}
    L_s(x,y) = \frac{1}{K}\big(\alpha \dot ([f(x)]_y - \beta)^2 + \sum_{i=1,i \neq y}^K ([f(x)]_i)^2 \big)
\end{align}
where $\alpha, \beta$ are hyper-parameters. This is the regular MSE loss when $\alpha=\beta=1$. Larger values of $\beta$ reduce soft-max entropy in the output by scaling positive output class signals while keeping negative ones close to zero. 

We use $\beta$ to control the interaction between component models. For example, when tasks are highly dissimilar, such as in task- and class-incremental continual learning, we use larger values of $\beta$ to reduce interference of component models trained on other tasks upon composition. On the other hand, when tasks are highly synergistic such as in data-incremental learning, lower values of $\beta$ can be used to encourage interaction in the final composed model. 

\begin{table}[t]
    \centering
    \caption{Data-Incremental Learning on D-MIT-67, 4 tasks using ResNet-18. For fair comparison, we run our method under the settings proposed by \cite{shon2022dlcft}. \method{} outperforms the best replay-free method \cite{li2017learning} by $4.61\%$, and the best replay-based method \cite{shon2022dlcft} by $1.27\%$, while supporting parallel training across tasks (PL). If we sacrifice parallelization by initializing each model at task $t$ with $h_{t-1}$ (\method{}-Seq), we further improve over \method{} by $0.28\%$.}
    \begin{tabular}{lccc}
    \toprule
    Method & PL? & Replay Buffer & Accuracy \\
    \midrule
    SGD  & \xmark & \textbf{None} & $63.48$ \\
    LwF\cite{li2017learning} & \xmark & \textbf{None} & $67.21$ \\
    MAS\cite{aljundi2018memory} & \xmark & \textbf{None} & $62.49$ \\
    OSLA\cite{ritter2018online} & \xmark & \textbf{None} & $64.08$ \\
    EWC \cite{kirkpatrick2017overcoming} & \xmark & 500 & $63.68$ \\
    DLCFT\cite{shon2022dlcft} & \xmark & 500 & $70.55$ \\
    \textbf{\method{}} & \cmark & \textbf{None} & $\bm{71.82}$ \\
    \textbf{\method{}-Seq} & \xmark & \textbf{None} & $\bm{72.10}$ \\
    \midrule
    Joint (Linear) & - & - & 73.61 \\
    Joint (Paragon) & - & - & 74.40\\
    \bottomrule
    \end{tabular}
    \label{tab:mit-nlcft-comparison}
    \vspace{-3mm}
\end{table}

\begin{table*}[t]
\centering
\caption{{\bf Task-Incremental Learning:} Comparison against existing methods using ImageNet pre-trained ResNet-50. We almost uniformly outperform all replay-based and replay-free methods that we compare against. In particular, we improve over the best replay-free (RF) method by $2.46\%$, and the best replay-based method by $1.38\%$ \cite{boschini2022transfer}.}
\begin{tabular}{lccc|ccc}
\toprule
Method / Buffer      & \multicolumn{3}{c}{C-Caltech-256} & \multicolumn{3}{c}{C-MIT-67} \\
 \cmidrule(lr){2-4} \cmidrule(lr){5-7}  
\multicolumn{1}{r}{Tasks}       & 5        & 10        & 20       & 5      & 10      & 20     \\
\midrule
ER \cite{robins1995catastrophic} / 300  & $92.11$\tiny{$\pm0.26$} & $94.33$\tiny{$\pm0.48$}  &   $95.76$\tiny{$\pm0.13$}  &  $85.84$\tiny{$\pm0.09$} & $90.59$\tiny{$\pm0.51$}  & $95.13$\tiny{$\pm0.27$} \\
LWF \cite{li2017learning} / \textbf{None} & $93.73$\tiny{$\pm0.12$} & $95.31$\tiny{$\pm0.08$} & $96.29$\tiny{$\pm0.05$} & $87.55$\tiny{$\pm0.20$} & $90.82$\tiny{$\pm0.04$} & $95.68$\tiny{$\pm0.58$}\\
oEWC \cite{schwarz2018progress} / \textbf{None} & $83.97$\tiny{$\pm0.71$} & $76.20$\tiny{$\pm16.07$} & $91.49$\tiny{$\pm0.30$} & $67.31$\tiny{$\pm3.22$} & $68.45$\tiny{$\pm4.93$} & $84.74$\tiny{$\pm1.96$} \\
DER++ \cite{buzzega2020dark} / 300 & $93.79$\tiny{$\pm0.13$} & $94.92$\tiny{$\pm0.11$} & $95.87$\tiny{$\pm0.17$} &  $86.83$\tiny{$\pm0.26$} &  $91.63$ \tiny{$\pm0.75$} & $94.54$\tiny{$\pm0.62$} \\
Co2L \cite{cha2021co2l} / 300   &  $34.51$\tiny{$\pm3.05$}        & $39.62$\tiny{$\pm1.70$}   &  $45.94$\tiny{$\pm1.47$}  & $39.32$\tiny{$\pm20.47$}  &  $64.40$\tiny{$\pm1.31$}    &  $51.50$\tiny{$\pm2.21$} \\
ER-ACE \cite{caccia2022new} / 300  & $93.11$\tiny{$\pm0.12$} &   $94.63$\tiny{$\pm0.06$}  & $95.86$\tiny{$\pm0.10$} &  $87.13$\tiny{$\pm0.16$}  & $90.96$\tiny{$\pm0.28$}  & $94.92$\tiny{$\pm0.60$} \\
TWF (RF) \cite{boschini2022transfer} / \textbf{None} &  $\bm{94.45}$\tiny{$\pm0.05$} & $95.64$\tiny{$\pm0.06$} &   $96.36$\tiny{$\pm0.12$} & $86.76$\tiny{$\pm0.25$} &  $88.27$ \tiny{$\pm0.54$}  &  $94.74$\tiny{$\pm0.35$} \\
TWF \cite{boschini2022transfer} / 300 &  $94.38$\tiny{$\pm0.08$} & $95.74$\tiny{$\pm0.17$} &   $96.59$\tiny{$\pm0.07$} & $88.22$\tiny{$\pm0.41$} &  $92.13$ \tiny{$\pm0.26$}  &  $95.67$\tiny{$\pm0.46$} \\
\method{} / \textbf{None}    &  $94.33$ \tiny{$\pm0.03$}        &           $\bm{96.03}$ \tiny{$\pm0.02$} &  $\bm{97.22}$ \tiny{$\pm 0.03$}   &  $\bm{91.27}$\tiny{$\pm0.21$}    & $\bm{94.45}$\tiny{$\pm0.13$}     &   $\bm{97.71}$ \tiny{$\pm0.13$}    \\
\bottomrule
\end{tabular}
\vspace{-3mm}
\label{tab:task-incremental-sota}
\end{table*}

\section{Experiments}
\label{sec:experiments}
We describe comparison baselines in \secref{sec:baselines}, and our main results in \secref{sec:main-results}, with additional studies in \secref{sec:further-studies}. 

Our method is replay-free. As such, its computational cost is at most half of what it would be if it used a replay buffer since replay-based methods typically sample separate batches of equal size from both the current task and replay buffer at each iteration \cite{buzzega2020dark,boschini2022transfer,cha2021co2l,robins1995catastrophic,caccia2022new}. Nonetheless, we compare our method to both replay-based and replay-free methods in our baselines. While replay-based methods have thus far outperformed replay-free ones, ours is uniformly more accurate than replay-based methods tested, in addition to replay-free ones. 

We evaluate our method on Caltech-256~\cite{griffin2007caltech}, MIT-67~\cite{quattoni2009recognizing}, and OxfordPets~\cite{parkhi2012cats}.
For continual fine-tuning, we further split each training dataset into multiple disjoint subsets, each representing a single task, for fair comparison across experiments. We prefix sequential datasets split into tasks with disjoint labels (for task- and class-incremental learning) with C-, and random splits using a standardized random seed (for data-incremental learning) with D-. We set $\alpha=1, \beta=25$ for the former, $\alpha=1, \beta=5$ for the latter, and $\alpha=1, \beta=5$ for \methodensemble{}. Results in all tables and figures are averaged across 3 runs.

\subsection{Composition Baselines}
\label{sec:baselines}

In each of the following sections, we compare against the following baselines for model aggregation:

\noindent\textbf{Logits Ensemble (Ens-L): } Averaging the output logits of non-linear component models.

\noindent\textbf{Softmax Ensemble (Ens-SM): } Averaging the softmax output of non-linear component models.

\noindent\textbf{Weight composition of non-linear models (Soup): } Averaging the weights of non-linear component models. This method is inspired by Model Soups \cite{wortsman2022model} where weights of various non-linear models trained on the same task but with different hyper-parameters and augmentations are averaged to improve generalization at no additional inference cost.

\begin{figure}[t]
    \centering
    \includegraphics[width=0.35\textwidth]{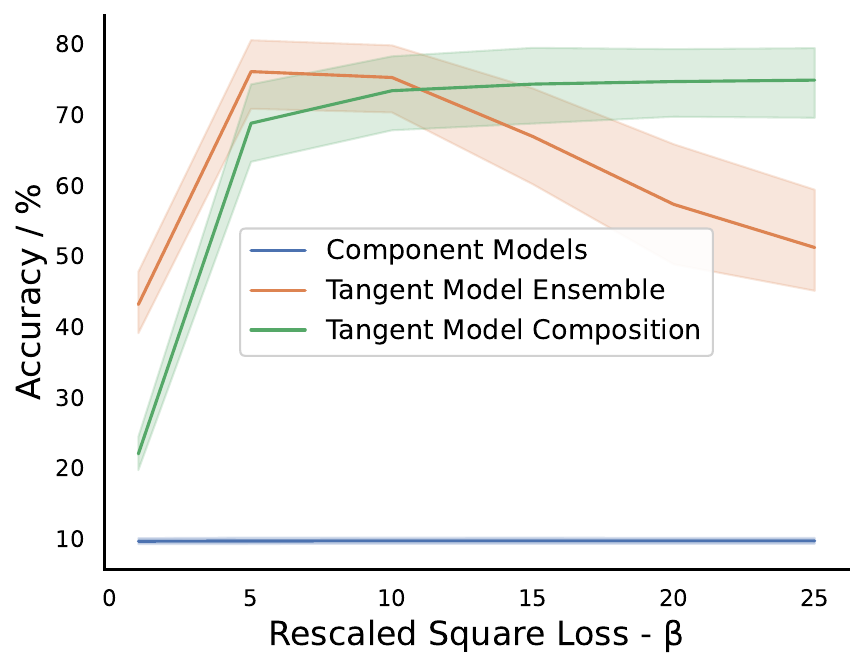}\\
    \includegraphics[width=0.35\textwidth]{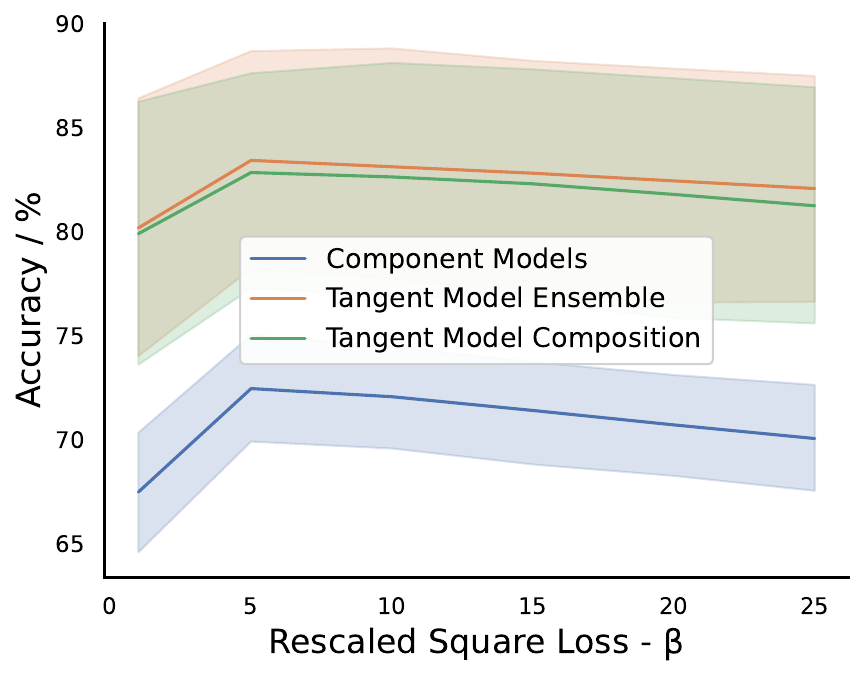}
    \caption{Ablation on $\beta$ averaged across \textbf{(Top:)} C-Caltech-256, C-MIT-67, and C-OxfordPets and \textbf{(Bottom:)} D-Caltech-256, D-MIT-67, and D-OxfordPets, split into 10 disjoint tasks. For class-incremental datasets, larger values of $\beta$ result in better generalization of the final composed model. 
    For data-incremental datasets where tasks are synergistic, larger values of $\beta$ instead harm the accuracy of both the component models and the composed model.
    }
    \label{fig:class-vs-data-incremental-rsl}
    \vspace{-3mm}
\end{figure}

\subsection{Main Results}
\label{sec:main-results}
We show that \method{} outperforms existing methods in all three continual fine-tuning settings \cite{van2019three}, from hardest to easiest: Class-Incremental (\secref{sec:class-incremental}), Data-Incremental (\secref{sec:data-incremental}), and Task-Incremental (\secref{sec:task-incremental}). 

\subsubsection{Class-Incremental Learning}
\label{sec:class-incremental}
In class-incremental learning, datasets are partitioned so that label spaces of each task are disjoint.

\noindent\textbf{Comparison against Composition Baselines: }
We compare against model composition baselines in \tabref{tab:continual-composition}. While \methodensemble{} yields the best generalization performance, this comes at a cost of $\mathcal{O}(T)$ inference time. On the other hand, \method{} achieves an $\mathcal{O}(1)$ inference time with only a 1.73\% trade-off in generalization accuracy on average, while outperforming weight averaging, logit ensembling, and even soft-max ensembling of non-linear models by an average of $7.83\%$, $11.37\%$, and $6.08\%$ respectively, even though the latter two both require $\mathcal{O}(T)$ inference time. 

\noindent\textbf{Comparison against existing methods: } In \tabref{tab:class-incremental-sota}, we compare against recent continual fine-tuning methods including both replay-free and replay-based methods. Here, replay-free methods \cite{li2017learning,schwarz2018progress} perform poorly in the class-incremental setting relative to stronger replay-based methods \cite{robins1995catastrophic,caccia2022new,buzzega2020dark,boschini2022transfer}. However, we show that not only do we improve over the best replay-free method by $14.48\%$, but also over the best replay-based method by $1.53\%$ \cite{boschini2022transfer}, corresponding to a  $17.59\%$ relative error reduction towards the paragon performance of joint training (\tabref{tab:continual-composition}).

\noindent\textbf{Comparison against EFCIL methods: } In \tabref{tab:efcil-tinyimagenet}, we compare against EFCIL methods on C-TinyImageNet \cite{le2015tiny}. The original EFCIL methods are evaluated under a different setting (denoted $\dagger$), where the zeroth task contains $50\%$ of the dataset. Hence, we further modified two of the best performing EFCIL methods, SSRE \cite{zhu2022self} and FeTrIL \cite{petit2023fetril} to use the pre-trained ImageNet initialization, and run them under our setting of uniformly sharded data. TMC not only outperforms all methods ran under our harder setting, by an average of $15.3\%$ over the next best method \cite{petit2023fetril}, but even yields equal or better performance when compared to all methods ran under $\dagger$.

\begin{table*}[h]
    \centering
    \caption{Comparison against EFCIL methods using either ImageNet pre-trained ResNet-50, or under the original EFCIL setting where the zeroth task contains $50\%$ of the dataset (denoted $\dagger$). All experiments are ran on C-TinyImageNet. \emph{Italics} denote the best method under $\dagger$, and \textbf{bold} denotes the best method under our harder setting. Under the former, TMC performs equally or better than all methods. Under the latter, TMC outperforms the best method by an average of $15.3\%$.}
    \begin{tabular}{lccccccc}
    \toprule
    Tasks & PASS$^\dagger$ \cite{zhu2021prototype} & IL2A$^\dagger$ \cite{zhu2021class} & SSRE$^\dagger$ \cite{zhu2022self} & FeTrIL$^\dagger$ \cite{petit2023fetril} & SSRE\cite{zhu2022self} & FeTrIL\cite{petit2023fetril} & \textbf{TMC} \\
    \cmidrule(lr){1-1} \cmidrule(lr){2-5} \cmidrule(lr){6-8}
    5 & 49.6 & 47.3 & 50.4 & \emph{54.8} & 33.7 & 44.1 & \textbf{57.6} \\
    10 & 47.3 & 44.7 & 48.9 & \emph{53.1} & 25.6 & 36.1 & \textbf{53.1} \\
        \bottomrule \\
    \end{tabular}
    \label{tab:efcil-tinyimagenet}
    \vspace{-7mm}
\end{table*}

\subsubsection{Data-Incremental Learning}
\label{sec:data-incremental}
Also referred to as domain-incremental learning, this setting assumes that tasks are split in a non-stratified manner, where each task has the same output space \cite{van2019three,shon2022dlcft}. 

\noindent\textbf{Comparison against Composition Baselines: }
In \tabref{tab:continual-composition}, we compare against model composition baselines. Similar to the class-incremental setting, \methodensemble{} outperforms \method{} by an average of $0.66\%$ at a cost of $\mathcal{O}(T)$ inference time. \method{} uses $\mathcal{O}(1)$ inference time and outperforms non-linear composition, logit ensemble, and even soft-max ensemble by an average of $7.04\%$, $6.32\%$, and $5.21\%$ respectively. 

\noindent\textbf{Comparison against existing methods: }
In \tabref{tab:mit-nlcft-comparison}, we benchmark our model against results obtained by \cite{shon2022dlcft} on a pre-trained ResNet-18 model using the D-MIT-67 dataset split into 4 disjoint subsets. We improve over the best replay-free method \cite{li2017learning} by $4.61\%$, and the best replay-based method \cite{shon2022dlcft} by $1.27\%$, corresponding to a $64\%$ and $33\%$ relative error reduction towards the joint training paragon respectively. \method{}-Seq, which uses sequential initialization by initializing each component model $\tilde{h}_t$ with $h_{t-1}$ instead of $h_0$, further improves over \method{} by $0.28\%$ but at the cost of parallelizability.

\subsubsection{Task-Incremental Learning}
\label{sec:task-incremental}
In task-incremental learning, the output spaces of different tasks are disjoint, but the source task of each sample is known at inference time. Thus, we can restrict the model predictions to that of the given task during evaluation.

 \noindent\textbf{Comparison against Composition Baselines: } Due to the disjoint output spaces of each task, ensembling outputs of specialist component models trained on individual tasks and restricting them to a target task essentially only considers the predictions of the specialist model trained on the target task (since, in general, specialist models are not effective discriminants for tasks other than those they are traiend on). Hence, the independence of output spaces renders ensembles of specialist models highly effective. In contrast, composing non-linear models in weight space does not ensure minimal interference across different tasks. Indeed, we show in \tabref{tab:continual-composition} that \methodensemble{} yields the best results, while the non-linear composition approach yields the worst in task-incremental learning. On the other hand, due to the equivalence of \method{} to (logit) ensembling of linearized models, \method{} outperforms not only weight averaging, but also logit and soft-max ensembles of non-linear models, by $3.78\%$, $1.83\%$, and $1.79\%$ respectively. 

\noindent\textbf{Comparison against existing methods: }
Since each component model is trained only on task $t$, by design, there is no interference from other tasks when training $\tilde{h}_t$. As such, the final composed model is equivalent to a specialist model when restricted to any task. However, in other state of the art continual learning methods such as those involving replay or loss regularization, this separation among trained for each task is violated. 
Hence, under the task-incremental setting, we show in \tabref{tab:task-incremental-sota} that \method{} achieves the best performance across both replay-free and replay-based methods even without using any memory buffer, beating the next best method by $2.46\%$ and $1.38\%$ respectively \cite{boschini2022transfer}.

\subsection{Further studies}
\label{sec:further-studies}
We present additional studies on the Rescaled Square Loss (\secref{sec:ablation-rsl}), and compare against \method{}-Seq to show that our method yields an effective and highly parallelizable training scheme for continual fine-tuning (\secref{sec:parallel}).

\subsubsection{Rescaled Square Loss}
\label{sec:ablation-rsl}
In \secref{sec:method-rsl}, we showed that $\beta$ can be used to control the amount of interaction between tasks. In \figref{fig:class-vs-data-incremental-rsl}, we empirically show that in the class-incremental setting where tasks are highly dissimilar, larger values of $\beta$ result in a better generalist model. On the other hand, under the data-incremental setting where tasks are more similar, information from other tasks can benefit the learning of any given task. Here, larger values of $\beta$  instead harm generalization due to lowering the accuracy of each component model.

\begin{figure}[h]
    \centering
    \includegraphics[width=0.85\linewidth]{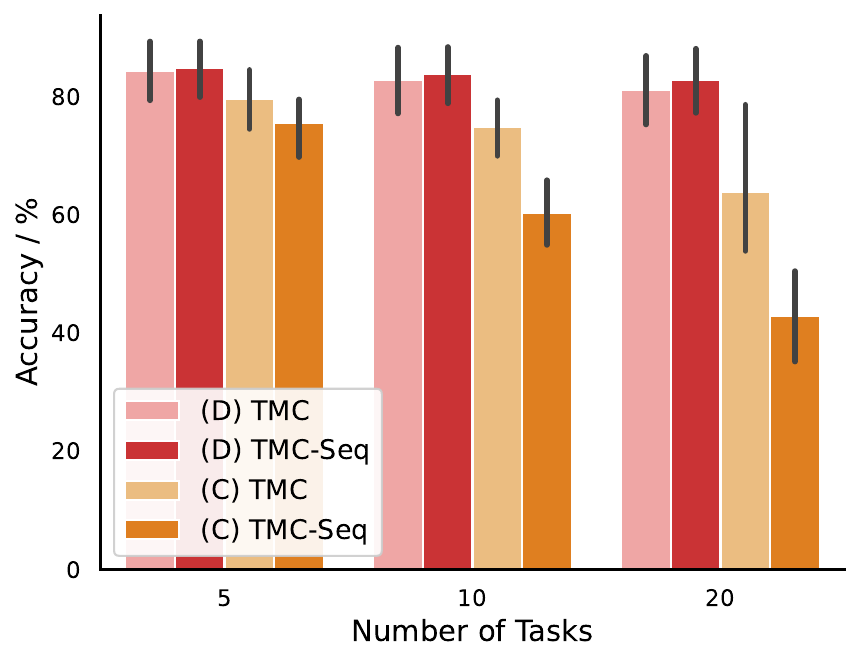}
    \caption{Accuracy of composed models, where component models are initialized with $h_0$ \textbf{(\method{})} and $h_{t-1} $\textbf{(\method{}-Seq)}. In the data-incremental setting \textbf{(D)}, {\method{}-Seq} outperforms \method{}, while in the class-incremental setting \textbf{(C)}, \method{} performs significantly better.  All results are averaged across D/C-Caltech-256, D/C-MIT-67, D/C-OxfordPets (we exclude C-OxfordPets in the 20-task setting since it does not contain enough classes).
    }
    \label{fig:seqcomp}
    \vspace{-5mm}
\end{figure}
\subsubsection{Convexity and Parallelizability}\
\label{sec:parallel}
Given a sequence of tasks $1 \ldots T$ and a pre-trained model $h_0$, there are at least two natural ways to learn the component model $\tilde{h}_{t}$: (1) Train $\tilde{h}_{t}$ with $h_{t-1}$ as initialization, and (2) Train $\tilde{h}_{t}$ with $h_0$ as initialization. The first option (\method{}-Seq) is most common among continual learning methods which impose regularization in the weights between the current and previous tasks or optimize using memory buffers consisting of examples obtained from previous tasks. As such, these methods are restricted to a sequential learning framework, and hence cannot be parallelized across tasks. In \secref{sec:method}, we showed  that, thanks to the convexity of the loss landscape, both options (1) and (2) are equivalent under our framework when we assume comparable generalization performance among global optima. 

In practice, for massively over-parameterized networks, such assumptions do not hold. However, we show empirically in \figref{fig:seqcomp} that under the data-incremental setting, compared to \method{}-Seq, \method{} only incurs a small loss in generalization performance to enable massive parallelization across tasks. This can potentially provide orders of magnitude speed-ups compared to existing continual learning methods. Moreover, under the class-incremental setting, \method{} greatly outperforms \method{}-Seq, showing that the pre-trained initialization yields solutions of better generalization than $h_{t-1}$. We attribute this to the fact that the hypothesis space of models which are globally optimal for the training data is vast and yields varying generalization. Thus, regardless of the loss landscape convexity, convergence to each point in this space depends on the ``direction of approach". Hence, initialization at $h_{t-1}$ introduces a bias towards the previous tasks. Due to the high level of dissimilarity between tasks in the class-incremental setting, this bias harms the final model generalization performance.

\section{Discussion}
\label{sec:discussion}
Our approach enables incremental learning under the assumption that increment tasks are ``close'' to the pre-trained model. We ensure proximity by training increment models independently on the tangent plane of the pre-trained model. 
The advantage is that pre-trained tasks are not forgotten, and increment tasks can be learned independently, in parallel, and easily forgotten if needed.

Of course, this approach does not address the problem of incremental learning in full generality, when increment tasks can be arbitrary and arbitrarily unrelated. Nonetheless, we have shown that our method is competitive with existing methods in some of the most challenging settings, for instance when the hypothesis spaces of component tasks are disjoint. On the positive side, despite not requiring a replay buffer, our method outperforms replay-based ones under the assumptions in which we operate.

A limitation of our approach is that it requires computing the linear span of pre-trained models, which can be challenging. However, for common deep network architectures such as the ResNet \cite{he2016deep} or Transformer \cite{vaswani2017attention,dosovitskiy2020image} family, the Jacobian-Vector product can be efficiently evaluated \cite{pearlmutter1994fast,mu2020gradients,achille2021lqf,liu2023tangent}, reducing the computational complexity of training the tangent model to that of the same magnitude as a linear classifier on network activations. Hence, inference time for the linearized network is at most double that of the original model.

Another limitation of our approach is assuming a first-order expansion around a pre-trained objective can well approximate the fine-tuned model. Such assumptions hold poorly when the pre-training and downstream tasks are highly unrelated. Nevertheless, an ImageNet \cite{deng2009imagenet} pre-trained model already yields a sufficiently good local approximation for various real-world datasets \cite{achille2021lqf,liu2023tangent}.

While we mainly explored linear compositionality in the context of continual learning, we note that this framework can be easily generalized. We discussed possible applications towards federated learning and forgetting, since our method naturally yields a parallel training framework that fully compartmentalizes information obtained each task.

\section{Acknowledgements}
We would like to thank Aditya Golatkar, Albert Zhao, and the anonymous reviewers for their feedback on the initial version of the paper. This work was supported by ARO W911NF-17-1-0304 and ONR N00014-22-1-2252.

{\small
\bibliographystyle{ieee_fullname}
\bibliography{refs}
}

\clearpage
\appendix
\onecolumn
\begin{center}
\Large\bfseries Supplementary Material
\end{center}

\section{Implementation Details}
We run all our experiments on a ResNet-50 model pre-trained with ImageNet, except in Table.~3 where we use a pre-trained ResNet-18 model instead for fair comparison with existing benchmarks. To construct the tangent models, we re-initialize the last fully connected layer such that the number of output classes matches that of the target tasks. We train the original models with SGD using the cross-entropy loss. Tangent models are trained with the Adam optimizer using the rescaled squared loss (RSL).  We train all (original and tangent) models for 50 epochs using a decaying learning rate factor of $0.1$ at epochs 25 and 40, and use a constant batch size of $32$ across all experiments. We average each experiment across 3 independent runs. 

For all baseline methods, we follow \cite{buzzega2020dark,buzzega2021rethinking,boschini2022transfer} and standardize batch size at 32. The only two exceptions are Co2L \cite{cha2021co2l}, where we set batch size and buffer batch size to 256 since a large batch size is needed to achieve reasonable accuracy, and TWF \cite{boschini2022transfer} where we set both task and buffer batch size to $12$ due to its larger memory requirement resulting from the attention mechanism. 
For replay-based baselines, at each iteration, in addition to sampling a batch from the current task, we also sample a batch of the same size (32) from the replay buffer. Note that this immediately increases training time for all replay-based methods by a factor of two, since the training set size for each task is effectively doubled for fixed number of epochs. 

We prepare our datasets as follows: 
\textbf{(a) Caltech-256~\cite{griffin2007caltech}} contains $\sim30.6K$ images across $256$ object classes. Following \cite{achille2021lqf}, we sample 60 images per class to train and test on the remaining.
\textbf{(b) MIT-67~\cite{quattoni2009recognizing}} contains $\sim15.6K$ images across $67$ indoor scene categories. We split MIT-67 using the provided train-test splits. \textbf{(c) OxfordPets~\cite{parkhi2012cats}} contains $\sim7.3K$ images across $37$ pet categories. We split the dataset equally for training and testing. 

We set $\lambda_{(t,1)} = \frac{1}{t}$ and $\lambda_{(t,2)} = \frac{t-1}{t}$, which weights each model by the number of component models used to construct it. This is a natural choice when tasks are assumed to be distributed uniformly. We note that in cases such as task imbalance, it is likely that there will be more effective choices for $\lambda$. We leave the investigation of this for future work.

\section{Extended Literature Review}
\begin{table}[h]
    \centering
    \caption{\textbf{Fine-tuning classification head:} Fine-tuning only the classification head and composing them is the simplest form of \method{} (we refer to this as \method{}-FC), where linearization is only done with respect to the last fully-connected classification layer. Here, we show that \method{}-FC is highly effective for C-Caltech-256 using an ImageNet pre-training objective, but \method{} works significantly better for C-MIT-67 and C-OxfordPets since the features learnt from ImageNet pre-training do not generalize as well to indoor scene recognition and fine-grained object (pet species) classification respectively.
  }
    \begin{tabular}{cccccccc}
    \toprule
     Dataset & Tasks & Soup & Ens-L  & Ens-SM & \method{}-FC & \method{} & \methodensemble{}  \\
     \multicolumn{2}{r}{Inference/Memory Cost:} & $\mathcal{O}(1)$ & $\mathcal{O}(T)$ & $\mathcal{O}(T)$ & $\mathcal{O}(1)$ & $\mathcal{O}(1)$ & $\mathcal{O}(T)$\\
         \midrule
\multirow{3}{*}{\shortstack{C-Caltech-256\\\small{\textit{(Class-IL)}}}} & 5  &  $84.00$ & $82.80$ & $83.09$ & $\bf{85.30}$ & $84.82$ & $\textit{85.53}$ \\
& 10 &  $80.10$ & $76.85$  & $79.76$ & $\bf{83.43}$ & $82.37$  & $\textit{82.58}$ \\
& 20 &  $74.86$ & $66.76$  & $75.47$ & $\bf{79.13}$ & $78.66$ & $\textit{79.30}$ \\
\hdashline
\multirow{3}{*}{\shortstack{C-MIT-67\\\small{\textit{(Class-IL)}}}} & 5 & $59.48$ & $56.79$ & $58.68$ & $61.50$ & $\bf{69.43}$ & $\textit{71.72}$ \\
& 10 &  $49.30$   & $42.36$ & $50.52$ & $54.53$ &  $\bf{64.53}$ & $\textit{65.72}$  \\
& 20 &  $28.51$  & $23.61$ & $35.62$ & $40.05$ & $\bf{48.98}$  & $\textit{54.93}$ \\
\hdashline
\multirow{2}{*}{\shortstack{C-OxfordPets\\\small{\textit{(Class-IL)}}}} & 5 & $80.71$  & $81.14$ &  $81.86$  &  $77.83$ &  $\bf{84.42}$ & $\textit{85.04}$ \\
& 10 &  $71.35$ & $69.71$ & $77.31$  & $70.93$ &  $\bf{77.75}$  & $\textit{79.97}$ \\
    \bottomrule
    \end{tabular}
    \label{tab:tmc-fc}
    \vspace{-3mm}
\end{table}

We describe in detail the three scenarios for continual learning proposed by \cite{van2019three}. In Task-Incremental Learning, task identity (Task-ID) is provided at inference time. In other words, models can be trained with task-specific components which are ``selected" during inference by the provided Task-ID. For example, in the case of multi-class classification, this corresponds to simply selecting the output nodes corresponding to the task, and restricting predictions only to this subset. Thus, preserving intra-task performance is critical for Task-Incremental Learning. In Data/Domain-Incremental Learning, Task-ID is not provided at inference time. However, knowledge of the task identity is not necessary for inference. \cite{van2019three} cites the example of protocols under which structure of each task remains consistent, but distribution of inputs differs across tasks. Lastly, Class-Incremental Learning requires both inferring Task-ID and solving the task at hand. For instance, this happens when each task contains a new subset of classes/labels that are not present in previously encountered tasks.

\section{Additional Discussion}

\subsection{Fine-tuning classification head}

The effectiveness of \method{} is partly due to the implicit regularization arising from linearization of a model about the pre-training objective. In the extreme case, we can choose only to linearize with respect to the last layer of a neural network, which in our scenario is the classification head. Note that this is often a fully-connected layer which is already linear. As such, this is equivalent to simply fine-tuning the classification head, which can be composed for inference. We refer to this as \method{}-FC, and show in \tabref{tab:tmc-fc} that this simplest form of linearization can yield even better accuracies than \method{} when the pre-training features are sufficient for the downstream task (in the case of C-Caltech-256), but performs significantly worse when this assumption does not hold (C-MIT-67 and C-OxfordPets).

\subsection{Failure modes of \method{}}
\begin{table}[t]
    \centering
    \caption{Non-linear fine-tuning on the DomainNet dataset.}
    \setlength{\tabcolsep}{3.5pt}
    \begin{tabular}{lccccccccc}
    \toprule
    \multirow{2}{*}{Test Domain} & \multicolumn{6}{c}{Train Domain} & \multicolumn{3}{c}{Composition} \\
    \cmidrule(lr){2-7} \cmidrule(lr){8-10}
    & Clipart &  Quickdraw & Photo & Infograph & Sketch & Painting & Ens-L & Ens-SM & Soup \\
    \midrule
    Clipart   &   $75.90 $\tiny{$\pm0.10 $} &    $4.56 $\tiny{$\pm1.42 $} &   $48.06 $\tiny{$\pm0.27 $} &   $38.32 $\tiny{$\pm0.26 $} &   $49.60 $\tiny{$\pm0.68 $} &   $42.57 $\tiny{$\pm0.91 $} &    $60.27 $\tiny{$\pm6.84 $} &   $70.68 $\tiny{$\pm0.13 $} &   $62.18 $\tiny{$\pm0.42 $} \\
    Quickdraw &    $9.45 $\tiny{$\pm0.25 $} &   $69.64 $\tiny{$\pm0.08 $} &    $4.85 $\tiny{$\pm0.13 $} &    $3.35 $\tiny{$\pm0.23 $} &   $10.56 $\tiny{$\pm0.16 $} &    $3.73 $\tiny{$\pm0.12 $} &    $25.11 $\tiny{$\pm0.48 $} &   $36.43 $\tiny{$\pm0.47 $} &   $14.30 $\tiny{$\pm0.21 $} \\
    Photo      &   $54.21 $\tiny{$\pm0.32 $} &    $1.68 $\tiny{$\pm0.38 $} &   $83.41 $\tiny{$\pm0.08 $} &   $53.27 $\tiny{$\pm0.45 $} &   $48.93 $\tiny{$\pm0.41 $} &   $60.20 $\tiny{$\pm0.49 $} &   $62.19 $\tiny{$\pm11.88 $} &   $77.51 $\tiny{$\pm0.23 $} &   $69.97 $\tiny{$\pm0.36 $} \\
    Infograph &   $18.47 $\tiny{$\pm0.24 $} &    $0.55 $\tiny{$\pm0.17 $} &   $21.84 $\tiny{$\pm0.11 $} &   $43.25 $\tiny{$\pm0.24 $} &   $15.33 $\tiny{$\pm0.16 $} &   $19.34 $\tiny{$\pm0.52 $} &    $24.05 $\tiny{$\pm3.96 $} &   $33.13 $\tiny{$\pm0.26 $} &   $26.04 $\tiny{$\pm0.14 $} \\
    Sketch    &   $41.40 $\tiny{$\pm0.77 $} &    $5.08 $\tiny{$\pm0.52 $} &   $37.00 $\tiny{$\pm0.20 $} &   $29.88 $\tiny{$\pm0.36 $} &   $69.14 $\tiny{$\pm0.16 $} &   $37.25 $\tiny{$\pm0.28 $} &    $54.54 $\tiny{$\pm2.86 $} &   $61.37 $\tiny{$\pm0.22 $} &   $52.88 $\tiny{$\pm0.42 $} \\
    Painting  &   $36.83 $\tiny{$\pm0.45 $} &    $0.70 $\tiny{$\pm0.11 $} &   $48.48 $\tiny{$\pm0.16 $} &   $35.70 $\tiny{$\pm0.34 $} &   $36.70 $\tiny{$\pm0.52 $} &   $71.53 $\tiny{$\pm0.33 $} &   $46.14 $\tiny{$\pm12.21 $} &   $63.64 $\tiny{$\pm0.54 $} &   $55.14 $\tiny{$\pm0.64 $} \\
    \midrule
    Average   &   $39.38 $\tiny{$\pm0.35 $} &   $13.70 $\tiny{$\pm0.45 $} &   $40.61 $\tiny{$\pm0.16 $} &   $33.96 $\tiny{$\pm0.31 $} &   $38.37 $\tiny{$\pm0.35 $} &   $39.10 $\tiny{$\pm0.44 $} &    $45.39 $\tiny{$\pm6.37 $} &   $57.13 $\tiny{$\pm0.31 $} &   $46.75 $\tiny{$\pm0.36 $} \\
    \bottomrule
    \end{tabular}
    \label{tab:domainnet-nonlinear}
\end{table}

\begin{table}[t]
    \centering
    \caption{Tangent Fine-tuning on the DomainNet dataset.}
    \setlength{\tabcolsep}{3.5pt}
    \begin{tabular}{lccccccccc}
    \toprule
    \multirow{2}{*}{Test Domain} & \multicolumn{6}{c}{Train Domain} & \multicolumn{3}{c}{Composition} \\
    \cmidrule(lr){2-7} \cmidrule(lr){8-10}
    & Clipart &  Quickdraw & Photo & Infograph & Sketch & Painting & \methodensemble{} & \method{} \\
    \midrule
    Clipart   &   $64.01 $\tiny{$\pm0.15 $} &    $3.52 $\tiny{$\pm0.09 $} &   $34.15 $\tiny{$\pm0.17 $} &   $24.43 $\tiny{$\pm0.23 $} &   $33.07 $\tiny{$\pm0.17 $} &   $27.70 $\tiny{$\pm0.09 $} &   $50.54 $\tiny{$\pm0.17 $} &   $57.10 $\tiny{$\pm0.23 $} \\
    Quickdraw &    $2.57 $\tiny{$\pm0.10 $} &   $56.70 $\tiny{$\pm0.07 $} &    $1.71 $\tiny{$\pm0.06 $} &    $1.15 $\tiny{$\pm0.03 $} &    $2.97 $\tiny{$\pm0.07 $} &    $1.19 $\tiny{$\pm0.07 $} &    $7.17 $\tiny{$\pm0.15 $} &   $25.54 $\tiny{$\pm0.63 $} \\
    Photo      &   $45.68 $\tiny{$\pm0.13 $} &    $2.71 $\tiny{$\pm0.02 $} &   $80.31 $\tiny{$\pm0.06 $} &   $47.08 $\tiny{$\pm0.04 $} &   $41.49 $\tiny{$\pm0.10 $} &   $54.18 $\tiny{$\pm0.06 $} &   $69.34 $\tiny{$\pm0.06 $} &   $74.50 $\tiny{$\pm0.05 $} \\
    Infograph &   $12.94 $\tiny{$\pm0.08 $} &    $0.75 $\tiny{$\pm0.04 $} &   $15.37 $\tiny{$\pm0.15 $} &   $33.66 $\tiny{$\pm0.10 $} &   $10.43 $\tiny{$\pm0.08 $} &   $12.89 $\tiny{$\pm0.09 $} &   $21.52 $\tiny{$\pm0.13 $} &   $23.63 $\tiny{$\pm0.15 $} \\
    Sketch    &   $25.24 $\tiny{$\pm0.12 $} &    $3.08 $\tiny{$\pm0.06 $} &   $23.10 $\tiny{$\pm0.02 $} &   $17.41 $\tiny{$\pm0.19 $} &   $54.65 $\tiny{$\pm0.13 $} &   $22.11 $\tiny{$\pm0.05 $} &   $37.43 $\tiny{$\pm0.11 $} &   $44.38 $\tiny{$\pm0.09 $}  \\
    Painting  &   $26.62 $\tiny{$\pm0.05 $} &    $1.68 $\tiny{$\pm0.01 $} &   $40.74 $\tiny{$\pm0.13 $} &   $27.97 $\tiny{$\pm0.12 $} &   $28.73 $\tiny{$\pm0.20 $} &   $63.26 $\tiny{$\pm0.06 $} &   $48.99 $\tiny{$\pm0.09 $} &   $55.36 $\tiny{$\pm0.03 $} \\
    \midrule
    Average   &   $29.51 $\tiny{$\pm0.10 $} &   $11.41 $\tiny{$\pm0.05 $} &   $32.56 $\tiny{$\pm0.10 $} &   $25.29 $\tiny{$\pm0.12 $} &   $28.56 $\tiny{$\pm0.12 $} &   $30.22 $\tiny{$\pm0.07 $} &   $39.17 $\tiny{$\pm0.12 $} &   $46.75 $\tiny{$\pm0.20 $} \\
    \bottomrule
    \end{tabular}
    \label{tab:domainnet-tmc}
\end{table}

As discussed in the main paper, while \method{} and \methodensemble{} is often more effective than weight averaging or ensembling of non-linear models partly due to the implicit regularities provided by model linearization, this can lead to over-regularization when the downstream task and pre-training objective are highly dissimilar. We demonstrate this using the DomainNet dataset consisting of 6 different domains - Cliparts, Google Quickdraws, Photos, Infographs, Sketches, and Paintings. We fine-tune, from an ImageNet pre-training initialization, individual non-linear and tangent models on each domain in \tabref{tab:domainnet-nonlinear} and \tabref{tab:domainnet-tmc} respectively. We show that in such cases, component models and the composed model trained using non-linear fine-tuning outperforms tangent fine-tuning, as a result of over-regularization in the tangent model.

\subsection{Visualizations of Component Model Accuracies}
\begin{figure}[h]
    \centering
    \includegraphics[width=0.85\linewidth]{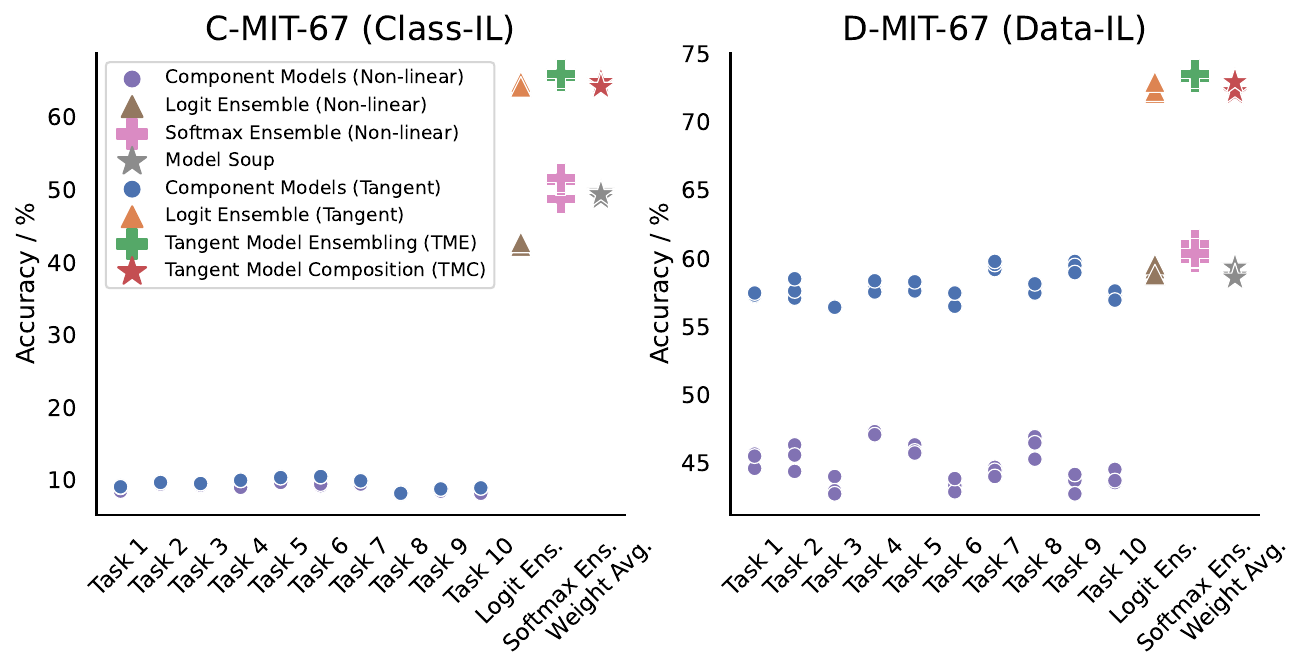}
    \caption{Accuracies of component and composed models for MIT-67 Dataset}
    \label{fig:task-breakdown-mit}
\end{figure}

\begin{figure}[h]
    \centering
    \includegraphics[width=0.85\linewidth]{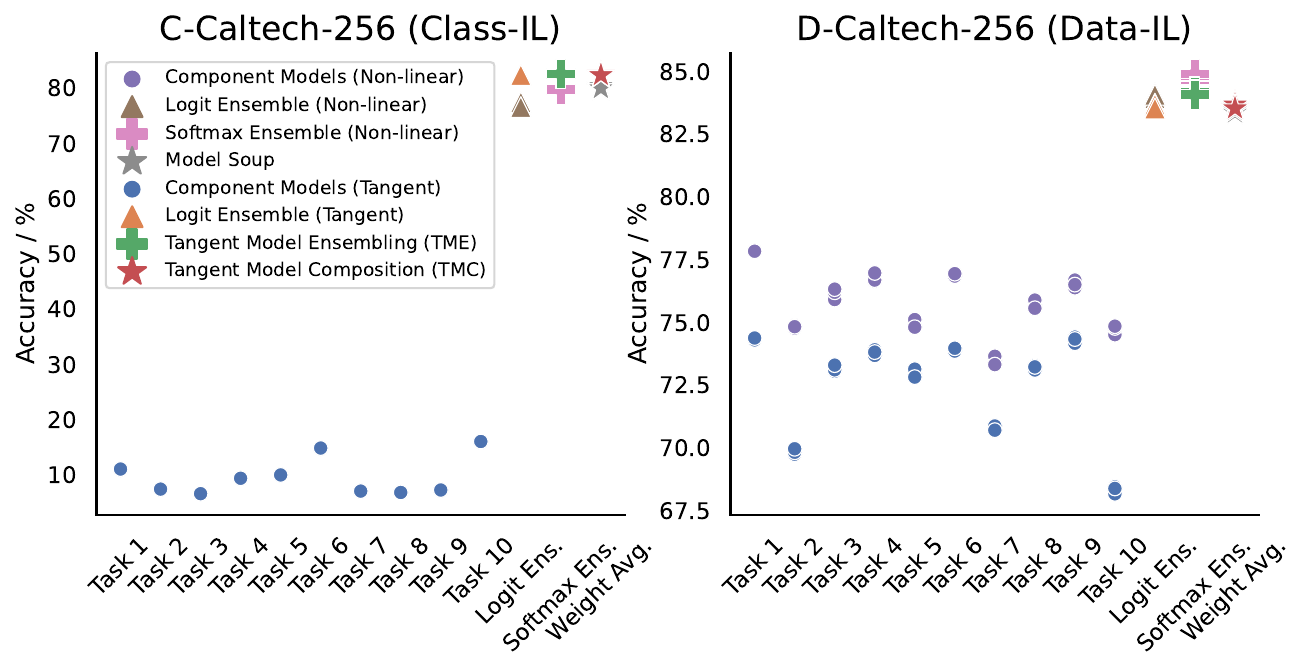}
    \caption{Accuracies of component and composed models for Caltech-256 Dataset}
    \label{fig:task-breakdown-caltech}
\end{figure}

\begin{figure}[h]
    \centering
    \includegraphics[width=0.85\linewidth]{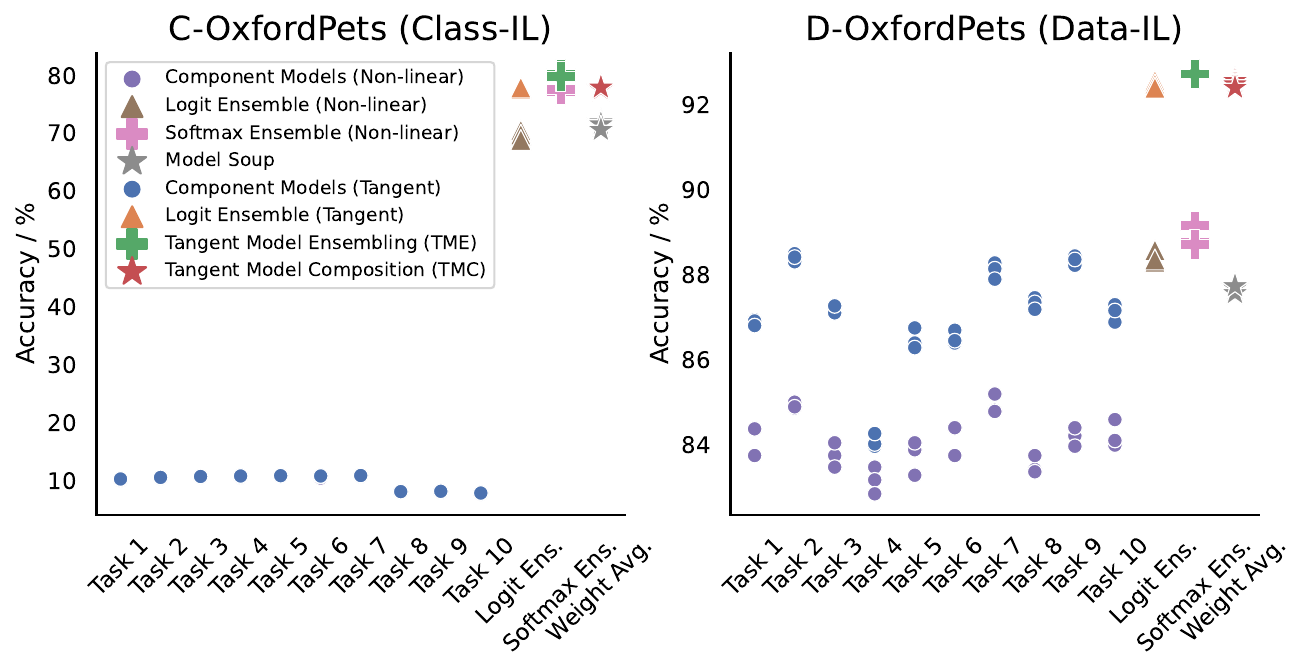}
    \caption{Accuracies of component and composed models for OxfordPets Dataset}
    \label{fig:task-breakdown-oxfordpets}
\end{figure}

We provide additional visualizations of the individual component model accuracies for MIT-67, Caltech-256, and OxfordPets in \figref{fig:task-breakdown-mit}, \figref{fig:task-breakdown-caltech}, and \figref{fig:task-breakdown-oxfordpets} respectively.


\subsection{Further Discussion on the Rescaled Square Loss}
Here, we elaborate on the effectiveness of RSL when tasks are highly dissimilar (class-incremental learning). First, note that due to random initialization, component models trained on tasks $\tau \neq t$ will produce non-zero noisy output signals for labels contained within task $t$ upon composition. While these noisy signals are too small to harm or bias the predictions of individual component models, the summation of the noise from each component model upon ensembling/composition can significantly affect the final prediction. 

\begin{figure}[h]
    \centering
    \includegraphics[width=1.0\linewidth]{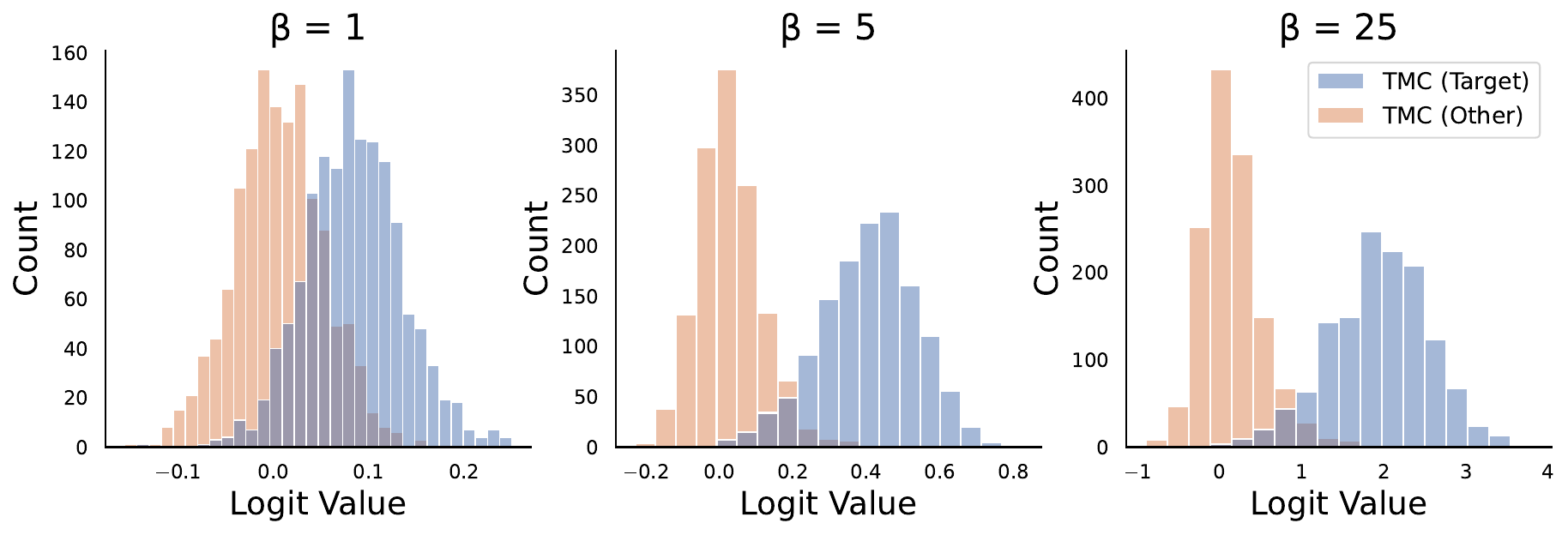}
    \caption{We plot the output logit distribution pertaining to the ground-truth class \textbf{[TMC (Target)]} and another class \textbf{[TMC (Other]}. Output values for the former should ideally be large, while output values for the latter should ideally be close to zero. We see that for small $\beta=1$, there is significant overlap between the two distributions, reducing the contrast between positive and negative output signals. On the other hand, larger values of $\beta$ produces significantly less overlap between the two distributions. Plots are done on C-MIT-67, 10 tasks.}
    \label{fig:rsl-beta-positive-negative}
\end{figure}

The effectiveness of soft-max ensembling demonstrated in the main paper can be attributed to the fact that the soft-max operation greatly increases the signal-to-noise ratio for each individual component model. This minimizes the sum of the noise components relative to the constructive signals present in the final composed model. Training using RSL with large values of $\beta$ aims to mimic this effect by encouraging greater separation between the positive and negative outputs, and thus increasing the signal-to-noise ratio of each component model.
We show the effects of $\beta$ on the output distribution of the composed model in \figref{fig:rsl-beta-positive-negative}. Lower values of $\beta$ reduces the signal-to-noise ratio, causing noisy signals to become more prominent in the final composed model. On the other hand, larger values of $\beta$ increases the distinction between the positive and negative signals in the final composed model.
On the other hand, in the case when tasks are similar, values of $\beta$ can result in ignoring weak but constructive signals from different component models. 

\end{document}